\documentclass[letterpaper, 10 pt, conference]{ieeeconf}  

\IEEEoverridecommandlockouts                              

\title{\LARGE \bf
Verification of Markov Decision Processes with Risk-Sensitive Measures
}

\author{Murat Cubuktepe and Ufuk Topcu
\thanks{Authors are with the Department of Aerospace Engineering and Engineering Mechanics, University of Texas, 201 E 24th St, Austin, TX 78712, USA.  e-mail: (\{mcubuktepe, utopcu\}@utexas.edu). The work has been supported partly by DARPA W911NF-16-1-0001, AFRL FA8650-15-C-2546 and ONR N00014-15-IP-00052.
}%
}
\newcounter{subeq}

\usepackage[utf8]{inputenc}
\usepackage{tabularx,ragged2e,booktabs}

\usepackage{subcaption}
\usepackage{amsmath,amssymb,amsfonts}
\usepackage{mathtools}
\usepackage[usenames,dvipsnames]{color}
\usepackage{xcolor}
\usepackage{xspace}
\usepackage{microtype}
\usepackage{todonotes}
\usepackage{colonequals}
 \usepackage{booktabs}

\usepackage{multirow}
\usepackage{multicol}

\usepackage{paralist}

\usepackage{url}
\usepackage{graphicx}
\usepackage{listings}
\usepackage{nicefrac}
\usepackage{tikz} 
\usepackage{pgfplots}

\usetikzlibrary{arrows,decorations.pathmorphing,positioning,fit,trees,shapes,shadows,automata,calc} 

\tikzset{outline/.style args={#1}{%
  draw=#1,thick,fill=#1!50}}

\newcommand{\dtmc}{\mathcal{D}}

\newcommand{\p}{\ensuremath{\mathbb{P}}}
\newcommand{\pr}{\ensuremath{\mathrm{Pr}}}
\newcommand{\reachPr}[2]{\ensuremath{\pr^{#1}(\finally #2)}}
\newcommand{\reachPrT}[1][]{\ensuremath{\reachPr{#1}{T}}}

\newcommand{\reachPrs}[3]{\ensuremath{\pr^{#1}_{#2}(\finally #3)}}

\newcommand{\finally}{\lozenge}

\newcommand{\sj}[1]{}

\newcommand{\R}{\mathbb{R}}

\newcommand{\Ireal}{[0,\, 1]\subseteq\mathbb{R}}  

\newcommand{\Distr}{\mathit{Distr}}

\newcommand{\distDom}{X}

\newcommand{\distFunc}{\mu}
\newcommand{\distDomElem}{x}

\newcommand{\Paramvar}{\ensuremath{{V}}\xspace}        

\newcommand{\sinit}{s_{\mathit{I}}} 
\newcommand{\mdp}{\mathcal{M}}
\newcommand{\MdpInit}[1][]{\ensuremath{\mdp{#1}=(S{#1},\sinit{#1},\Act,\probmdp{#1})}}
\newcommand{\pMdpInit}[1][]{\ensuremath{\mdp{#1}=(S{#1},\,\sinit{#1},\Act,\Paramvar,\probmdp{#1})}}
\newcommand{\probmdp}{\mathcal{P}}

\newcommand{\posy}{\ensuremath{f}}
\newcommand{\mono}{\ensuremath{g}}

\newcommand{\sched}{\ensuremath{\sigma}}

\newcommand{\Pol}{\ensuremath{\mathit{Pol}}}

\newcommand{\Act}{\ensuremath{\mathit{Act}}}
\newcommand{\act}{\ensuremath{\alpha}}
\newcommand{\pmdp}{\ensuremath{\mathcal{P}}}

\DeclareMathAlphabet{\mathpzc}{OT1}{pzc}{m}{it}
\def\presuper#1#2%
  {\mathop{}%
   \mathopen{\vphantom{#2}}^{#1}%
   \kern-\scriptspace%
   #2}

\usepackage{graphicx} 
\graphicspath{ \linespread{2}
{images/} }
\usepackage[ruled,vlined]{algorithm2e}

\newcommand{\RR}{\mathbb{R}}

\newtheorem{definition}{Definition}
\newtheorem{remark}{Remark}
\newtheorem{example}{Example}
\usepackage{color}
\DeclareMathOperator*{\maximize}{\textnormal{maximize}}

\pgfplotsset{compat=1.13}
\usepackage{multirow}
\usepackage[normalem]{ulem}
\useunder{\uline}{\ul}{}
\begin{document}
\maketitle
\begin{abstract}

We develop a method for computing policies in Markov decision processes with risk-sensitive measures subject to temporal logic constraints. Specifically, we use a particular risk-sensitive measure from cumulative prospect theory, which has been previously adopted in psychology and economics. The nonlinear transformation of the probabilities and utility functions yields a nonlinear programming problem, which makes computation of optimal policies typically challenging. We show that this nonlinear weighting function can be accurately approximated by the difference of two convex functions. This observation enables efficient policy computation using convex-concave programming. We demonstrate the effectiveness of the approach on several scenarios.

\end{abstract}

\section{Introduction}

Markov decision processes (MDPs) model sequential decision-making problems in stochastic dynamic environments \cite{puterman2014markov}. MDP formulations typically focus on the \emph{risk-neutral} expected cost or reward model. On the other hand, MDPs with \emph{risk-sensitive} measures, such as exponential utility \cite{howard1972risk}, percentile risk criteria \cite{filar1995percentile} and conditional value at risk \cite{chow2015risk,chow2014algorithms,tamar2015optimizing}. have been studied in the literature. MDPs also found applications in portfolio management \cite{bielecki1999risk}, robotics \cite{ono2015chance}, stochastic shortest-path problems \cite{bertsekas1991analysis}, optimal control \cite{hernandez1996risk} and operations research \cite{borkar2002q,howard1972risk}. These measures capture the variability in the cost due to stochastic transitions in an MDP, and aim to minimize  the effect of the outcomes with high cost. 

We focus on a particular risk-sensitive measure that comes from \emph{cumulative prospect theory} (CPT) \cite{tversky1992advances}. This measure is widely used in psychology and economics to build models that explain the risk-sensitive behavior of humans in decision-making. Empirical evidence suggest CPT characterizes human preferences in decision-making \cite{lopes1999role,tversky1992advances}. The key elements of this theory are a value function that is concave for gains, convex for losses, and steeper for losses than for gains, and a \emph{nonlinear} transformation of the probability range, which inflates small
probabilities and deflates high probabilities. It is also a generalization of other risk-sensitive measures like VaR or CVaR \cite{prashanth2016cumulative}. Additionally, with different nonlinear weighting functions, CPT-based measures can represent risk-taking measures as well as risk-averse measures.


We investigate \emph{model checking} with respect to \emph{temporal logic specifications}. \emph{Formal verification} of temporal logic specifications has been extensively studied for MDPs
with risk-neutral measures \cite{BK08}, and mature tools exist for efficient verification with such risk-neutral measures \cite{KNP11}. Probabilistic model checking verifies reachability properties such as “the probability of reaching a set of unsafe states is less than $5 \% $” and expected costs properties such as “the expected cost of reaching a goal state is less than $10 \% $”. A rich set of properties, specified by temporal logic specifications, can be reduced to \emph{reachability} properties, which can then be verified automatically \cite{katoen2016probabilistic}. To the best of our knowledge, formal quantitative verification with respect to risk-sensitive measures has not been considered in the literature. 


Dynamic programming equations for MDPs with CPT-based measures for finite-horizon MDPs in \cite{lin2013dynamic} and for infinite-horizon MDPs in \cite{lin2013stochastic} exist. However, computing an optimal policy requires optimizing integrals of nonlinear functions over continuous variables, which can be computationally impractical. CPT-based measures have been used in reinforcement learning \cite{prashanth2016cumulative}, where it was shown that the policy gradient approach converges to the optimal CPT value asymptotically.

The main challenge in computing policies with CPT-based measures is the nonlinear transformation of the probability range and utilities. This transformation yields a nonlinear programming problem. For efficient verification of MDPs with CPT-based measures, we approximate the nonlinear CPT weighting function by a \emph{difference of convex} function to utilize \emph{convex-concave procedure}  \cite{lipp2016variations}, which efficiently computes locally optimal solutions for optimization problems with difference of convex functions. We propose methods to approximate the CPT weighting function, and discuss the trade-offs between different approximations. Experimental results show the applicability of our approach in numerical experiments.


\section{Preliminaries}

\begin{definition}[Distribution] A \emph{probability distribution} over a finite or countably infinite set $\distDom$ is a function $\distFunc\colon\distDom\rightarrow\Ireal$ with $\sum_{\distDomElem\in\distDom}\distFunc(\distDomElem)=1$. 
The set of all distributions on $\distDom$ is denoted by $\Distr(\distDom)$.
\end{definition}

\begin{definition}[Monomials, Posynomials]\label{def:posy}
  Let $V=\{x_1,\ldots,x_n\}$ be a finite set of strictly positive real-valued \emph{variables}.
  A \emph{monomial} over $V$ is an expression of the form
  \begin{align*}
      \mono=c\cdot x_{1}^{a_{1}}\cdots x_{n}^{a_{n}}\ ,
  \end{align*} 
  
\noindent where $c\in \R $ is a real coefficient, and $a_i\in\R$ are exponents for $1\leq i\leq n$. 
  A \emph{posynomial} over $V$ is a sum of one or more monomials:
 

  \begin{align}
      \posy=\sum_{k=1}^K c_k\cdot x_1^{a_{1k}}\cdots x_n^{a_{nk}} \ .\label{eq:signomial}
  \end{align}
 \end{definition}
\begin{definition}[Markov decision process]\label{def:mdp}
A \emph{Markov decision process (MDP)} is a tuple $\pMdpInit$ with a finite set $S$ of states, an initial state $\sinit \in S$, a finite set $\Act$ of actions, and a transition function $\probmdp \colon S \times \Act \times S \rightarrow \Distr(S)$ satisfying for all $s\in S\colon \Act(s) \neq \emptyset$,  where $\Act(s) = \{\act \in \Act \mid \exists s'\in S.\,\probmdp(s,\,\act,\,s') \neq 0\}$. For given a state $s$, we denote the set of \emph{successor} states by $S(s)$. A state $s'$ is in $S(s)$ if there exists an $\act \in \Act$ such that $\probmdp(s,\,\act,\,s')>0$. If for all $s\in S$ it holds that $|\Act(s)| = 1$, $\mdp$ is called a \emph{discrete-time Markov chain (MC)}.
\end{definition}
%
%
%

$\Act(s)$ is the set of \emph{enabled} actions at state $s$; as $\Act(s) \neq \emptyset$, there are no deadlock states. \emph{Costs} are defined using a state--action \emph{cost function} $\mathit{C_t} \colon S \times \Act \times T \rightarrow \R_+$. \emph{Rewards} are defined similarly.

%

\begin{definition}[Policy]\label{def:scheduler}
	Given a finite horizon $T$, a (randomized) \emph{policy} for an MDP $\mdp$ is a function $\sched\colon S \times  T\rightarrow\Distr(\Act)$ such that $\sigma(s_t,\alpha) > 0$ implies $\alpha \in \Act(s_t)$ at time $t$. The set of memoryless policies over $\mdp$ at time $t$ is denoted by $\Pol_t^\mdp$, which only depends on the current state.
\end{definition}

\begin{definition}[Induced Markov chain]\label{def:induced_dtmc}
	For MDP $\MdpInit$ and policy $\sched\in\Pol^\mdp$, the \emph{Markov chain induced by $\mdp$ and $\sched$} is  $\mdp^\sched=(S,\sinit,\Act,\pmdp^\sched)$ where for all $s,s'\in S$,
	\begin{align*}
		 \pmdp^\sched(s,s')=\sum_{\alpha\in\Act(s)} \sched(s)(\act)\cdot\pmdp(s,\alpha,s').
	\end{align*} 
	
\end{definition}

We consider \emph{reachability properties}. For Markov chain $\dtmc$ with states $S$, let $\reachPrs{\dtmc}{s}{T}$ denote the probability of reaching a set $T \subseteq S$ of \emph{target states}  from  state $s\in S$; simply, $\reachPrT[\dtmc]$ denotes the probability for initial state $\sinit$. We use the standard probability measure as in~\cite{BK08}. 
The interest of this paper is a \emph{synthesis problem}, where the objective is to find a policy in $\Pol_t^\mdp$ such that the probability $\reachPrT[\dtmc]$ of satisfying the reachability property is maximized or minimized. 

The classical risk-neutral MDP problem is \cite{puterman2014markov}

\begin{align}
\inf_{\pi \in \Pol^\mdp}\mathbb{E}\left[\sum_{t=0}^{T}C_t(s_t,a_t)\right]. \label{eq:risk_neutral}
\end{align}

The problem in \eqref{eq:risk_neutral} can be solved with value iteration, policy iteration or linear programming, and the optimal policy will be a deterministic memoryless policy. The optimal policy for problem \eqref{eq:risk_neutral} will maximize the probability of satisfying the reachability property or minimize the expected cost, therefore it is a risk-neutral solution. Following from \cite{ruszczynski2010risk}, we consider the risk-sensitive value function starting at $s_0$, with a policy  $Pol^\mdp$, and the resulting trajectory ($s_0,\Pol_0^\mdp,s_1,\Pol_1^\mdp,...,s_T$), which is given by $C_T(\Pol^\mdp,s_0)= \rho_0(c(s_0,\Pol_0^\mdp)+\rho_1(c(s_1,\Pol_1^\mdp)+...+ \rho_{T-1}(c(s_{T-1},\Pol_{T-1}^\mdp)+C_T(s_T))...)),$ where $\rho_t$ is a one-step conditional risk measure at time $t$. Then, we consider the following optimization problem where $\rho$ is replaced by a CPT-based measure:

\begin{align}
\inf_{\pi \in \Pol^\mdp}C_T(\Pol^\mdp,\sinit). \label{eq:risk_CPT}
\end{align}

A dynamic programming equation exists for the problem in \eqref{eq:risk_CPT}, and the optimal policies are memoryless \cite{ruszczynski2010risk}. Any CPT-based measure is a one-step conditional risk measure, therefore the problem \eqref{eq:risk_CPT} can be solved by solving the dynamic programming equations \cite{lin2013dynamic}.
\section{Cumulative Prospect Theory (CPT)}
For a random variable $X$, the CPT \emph{value} is a generalization of the expected value of $X$ with a utility function that is concave on gains and convex on losses, and a probability weighting function that transforms the probability measure such that it inflates small probabilities and reduces larger probabilities.

\begin{definition}[CPT value]
For a random variable $X$, the CPT value is defined as

\begin{align}
C(X)= \int_0^{\infty} & w_{+}\left(\mathbb{P}\left(u_{+}\left(X\right)>z\right)\right) dz \nonumber \\
&-\int_0^{\infty} w_{-}\left(\mathbb{P}\left(u_{-}\left(X\right)>z\right)\right) dz,
\end{align}

\noindent where $w_{+}$ and $w_{-}$ $:$ $\left[0,1 \right]$ $\rightarrow$ $\left[0,1 \right]$ are two continuous non-decreasing functions with $w_{+}(0)=w_{-}(0)=0$ and $w_{+}(1)=w_{-}(1)=1$, $u_{+}$ and $u_{-}$ : $\mathbb{R}$ $\rightarrow$ $\RR_+$ are two utility functions.
\end{definition}

\begin{remark}
CPT value generalizes the expected value of a random variable, i.e, $C(X) = \mathbb{E}\left[X\right]= \int_0^{\infty} \left(\mathbb{P}\left( X>z\right)\right) dz-\int_0^{\infty} \left(\mathbb{P}\left( -X>z\right)\right) dz$, when $u_{+}(x)=u_{-}(x)=x$, and $w_{+}(x)=w_{-}(x)=x$.
\end{remark}

The functions $w_{+}$ and $w_{-}$ are the weighting functions that capture the concept of humans deflating high probabilities and inflating low probabilities when they make decisions under uncertainty. For instance, consider a scenario where one can earn \$100 with probability 1/100 and nothing otherwise, or can earn \$1 with probability 1. It is shown that the humans tend to choose the former option \cite{tversky1992advances,barberis2013thirty}, showing that the value of a decision by a human is nonlinear with respect to the transition probabilities. Reference \cite{prelec1998probability} suggests the weighting function $w(k)=\exp(-0.5(-\ln k)^\eta)$, with $0 < \eta < 1$ and \cite{tversky1992advances} suggests
\begin{align*}
w(k)=\dfrac{k^\eta}{\left(k^\eta+(1-k)^\eta\right)^{1 / \eta}}.
\end{align*}

Both of the functions have a similar inverted-S shape and they are concave for small values of $p$, and convex for large values of $p$.

The utility functions $u_{+}$ and $u_{-}$ represent how humans value gains $(X\geq 0)$ and losses $(X \leq 0)$ separately. For example, if we change the scenario in the above paragraph into losses, i.e, one will lose \$100 with probability 1/100 and nothing otherwise, or will lose \$1 with probability 1, then the humans tend to choose the latter option, showing that there is a difference between evaluating the gains and losses, and the CPT-based measures can handle losses and gains separately. A suggestion for the utility function is given in \cite{tversky1992advances}, which is $u_{+}(x)=\vert x \vert^{m}$, and $u_{-}(x)=-2.25\vert x\vert^{m}$, with $m=0.88$. Note that, $u_{+}$ is a concave function for $x>0$, and $u_{-}$ is a convex function for $x<0$.

\begin{remark}[\cite{prashanth2016cumulative}]
CPT-based measures generalize other risk-sensitive measures. For example, it is possible to represent value at risk or conditional value at risk by proper choice of weighting and utility functions.
\end{remark} 

\section{MDPs with CPT-based measures}
Reference \cite{lin2013dynamic} shows the existence of a dynamic programming equation in an MDP with CPT-based measures, and the optimal policy that comes from the dynamic programming equation is a memoryless randomized policy. Dynamic programming equations can be solved as a nonlinear programming problem. Specifically, the objective is a nonlinear function and the objective is  minimized or maximized  over randomized policies for a given state and time. However, solving optimization problems with a nonlinear objective function is generally impractical \cite{lin2013dynamic}.

%
%
%
%

To come up with a scalable procedure, we approximate the weighting function by a function that is the difference of two convex functions, which will reformulate the nonlinear programming problem to a difference of convex problem. Methods such as branch and bound methods \cite{lawler1966branch} or cutting plane methods \cite{androulakis1995alphabb} can find the globally optimal solution for a difference of convex problem, but these methods can be slow in practice. Instead of seeking a global solution, a locally optimal (approximate) solution can be found by utilizing the techniques of general nonlinear optimization \cite{nocedal2006sequential}.

\begin{definition}[Difference of convex problem]
Difference of convex (DC) problems have the following form
\end{definition}
\begin{align*}
			&\text{minimize } \quad f_{0}(x)-g_{0}(x)\\
			&\text{subject to}\quad f_{i}(x)-g_{i}(x)\leq 0, \quad i=1,2,...,m, \\
		\end{align*}


\noindent where $x \in \mathbb{R}^n$ is the variable vector, and the functions $f_i, g_i : \mathbb{R}^n \rightarrow \mathbb{R}$ for $i=0,1,...,m$ are convex.

The convex-concave procedure (CCP) \cite{lipp2016variations,shen2016disciplined} is a heuristic algorithm for finding a locally optimal solution to a DC problem. As a first step, we replace concave functions with a convex upper bound. We then solve the approximate convex problem, and the optimal value of the approximate problem will be an upper bound of the original problem at each iteration. The CCP algorithm to solve DC problems is described in Algorithm 1.
\begin{algorithm}[htb!]
\DontPrintSemicolon
\textbf{given} an initial feasible point $x_0$ and convex functions $f_i$, $g_i$.\;
k=0\;
\textbf{repeat}\;
\qquad 1. \textit{Convexify.} $\hat{g}_i(x)=g_i(x_k)+\nabla g_i(x_k)^T (x-x_k)$ for i=0,1,...,m\;
\qquad 2. \textit{Solve.} Set the value of $x_{k+1}$ to the solution of the convex problem\;
\qquad \qquad minimize $f_0(x)-\hat{g}_0(x)$\;
\qquad \qquad subject to $f_i(x)-\hat{g}_i(x)\leq 0$, \quad for i=1,2,...,m.\;
\qquad 3. \textit{Update iteration.} $k=k+1$,\;
\textbf{until} stopping criterion is satisfied.\;
\caption{CCP algorithm\label{IR}}
\end{algorithm}

Given an initial feasible point for a DC problem, (e.g. any policy from $\Distr(\Act)$), all of the successive iterates in Algorithm 1 will be feasible. The procedure given by Algorithm 1 is a descent algorithm, i.e, the objective will monotonically decrease over the iterations for a minimization problem or it will increase for a maximization problem, and it will converge to a local optimum \cite{lipp2016variations}. Therefore, the above algorithm can be used to compute locally optimal solutions by solving a sequence of convex optimization problems, which is efficiently solvable by well-studied methods \cite{boyd_convex_optimization}.

\subsection{Approximating the weighting function with a DC function}
In general, CPT weighting functions are nonlinear functions, and we can not use the weighting functions directly in convex-concave programming. Therefore, we approximate the weighting functions by a DC function to utilize convex-concave programming. A possible way to approximate the weighting function is least-squares polynomial approximation~\cite{de1978practical} or Chebyshev polynomial approximation~\cite{mason2002chebyshev}, but these methods can be inaccurate, as the CPT weighting functions that are frequently used in the literature are not Lipschitz continuous around zero probability. See Figure 1 for an example where the Chebyshev approximation method fails to approximate a weighting function.

\begin{figure}[ht]

\centering
  \input{myfile1.tex}
   \caption{An example of a CPT weighting function (red) and approximation of the CPT weighting function (blue) by a 25th degree Chebyshev polynomial with error tolerance $\epsilon=10^{-4}$. As the CPT weighting function is not Lipschitz, the approximation with a Chebyshev basis diverges with smaller $\epsilon$ with larger values of $p$.}
\end{figure}


Since the Chebyshev and least-squares polynomial methods perform poorly, we modify the least-squares polynomial approximation method by extending the polynomial basis functions with monomial basis functions to accurately approximate the CPT weighting function. For example, we approximate the function $\exp(-0.5(-\ln(k))^{0.9})$, which is used in \cite{lin2013dynamic}, by a posynomial function, $0.00231k^{0.05}+0.00128k^{0.1}+0.19578k^{0.35}+0.59897k^{0.4}+0.15968k^{0.95}+0.03318k^3+0.00847k^{23}.$ Figure 2 shows the posynomial function and the CPT weighting function.

\begin{figure}[ht]
\centering 
\begin{tikzpicture}
	\begin{axis}[xmin=0,xmax=1,ymin=0,ymax=1,mark=none,width=8.5cm,height=4.0cm,
		xlabel={Probability $k$},
		ylabel={$w(k)$},
        x label style={font=\normalsize},
        y label style={font=\normalsize},
	]
    \node[] at (40,90) {$\exp(-0.5(-\ln(k))^{0.9})$};
        \draw [->](38,81) -- (48,70);
	\addplot [domain=0:1, red, samples=700,mark=none]{0.00231642258521069*x^0.05+0.00128356642708694*x^0.1+0.195783466331253*x^0.35+0.598977890286512*x^0.4+0.159689481206954*x^0.95+0.0331820175871778*x^3+0.00847475103416698*x^23};
   \addplot [domain=0:1, blue,samples=700,mark=none]{e^(-0.5*(-ln(x))^0.9)}; 
   \addplot [domain=0:1, green,samples=10,mark=none]{x}; 
	\end{axis}
    
\end{tikzpicture}

\begin{tikzpicture}
	\begin{axis}[xmin=0,xmax=1,mark=none,width=8.5cm,height=4.0cm,
		xlabel={Probability $k$},
		ylabel={Approximation error},
        x label style={font=\normalsize},
        y label style={font=\normalsize},
	]


	\addplot [domain=0:1, red, samples=700,mark=none]{0.00231642258521069*x^0.05+0.00128356642708694*x^0.1+0.195783466331253*x^0.35+0.598977890286512*x^0.4+0.159689481206954*x^0.95+0.0331820175871778*x^3+0.00847475103416698*x^23-e^(-0.5*(-ln(x))^0.9)};

	\end{axis}
\end{tikzpicture}

   \caption{Top: An example of a CPT weighting function (blue) versus a regular transition function (green) and approximation of the CPT weighting function (red) by a DC posynomial function. Note that curve of the approximation is not visible, which shows that the approximation is accurate. Bottom: The error of the approximation of the CPT weighting function by a DC posynomial function.}
\end{figure}


\subsection{Computing locally optimal policies}

When the weighting functions are given as  $w_{+}(x)=w_{-}(x)=x$ and similarly for utility functions $u_{+}(x)=u_{-}(x)=x$, the dynamic programming equation to find the policy that maximizes the probability of satisfying the reachability property is
\begin{align}
&p_t(s)=\max_{\alpha \in Act}\sum_{s' \in S}\probmdp(s,\act,s') \cdot p_{t+1}(s')\nonumber, \\
&p_T(s)=1, \quad \forall s \in Q, \quad t=1,...,T, \label{eq:CPT_eq_reach}
\end{align}
where $p_t(s)$ denotes the probability of satisfying the reachability property at state $s$ and time $t$. Equivalently, we can write the dynamic programming equation in following for a given state $s$ and time $t$:
\begin{align}
&\maximize\limits_{\sched(s,a), a \in \Act} 
\sum_{\act\in\Act(s)}\sum_{s' \in S}\sched(s,\act)\cdot \probmdp(s,\act,s') \cdot p_{t+1}(s') \nonumber \\
&\enskip\text{subject to}\nonumber\\
&\sum_{\act\in\Act(s)}\sched(s,\act)=1, \quad \forall\act\in\Act(s), \quad \sched(s,\act) \geq 0.\label{eq:CPT_eq_reach1} 
\end{align}

The optimization problem in \eqref{eq:CPT_eq_reach1} maximizes the expected value of the probability for satisfying the reachability property, therefore it is a risk-neutral solution. Note that we can compute the expected value by solving the following problem:
\begin{align}
&\maximize\limits_{\sched(s,a), a \in \Act}  \sum_{s'_{q=1} \in S(s)}^{s'_{|S(s)|}\in S(s)}\bigg(\Phi_{q}\cdot
\Big(p_{t+1}(s'_{q})-p_{t+1}(s'_{q-1})\Big)\bigg)\nonumber\\ 
&\enskip\text{subject to}\nonumber\\ 
&\sum_{\act\in\Act(s)}\sched(s,\act)=1,\quad\forall\act\in\Act(s), \quad\sched(s,\act) \geq 0,\nonumber\\
&\Phi_{q} =\sum_{\act\in\Act(s)}\sum_{s'_{m=q} \in S(s)}^{s'_{|S(s)|} \in {S(s)}}\sched(s,\act)\cdot\probmdp(s,\act,s'_{m}), \label{eq:CPT_eq_reach2}
\end{align}
where $q=1,2,\ldots,|S(s)|$ gives the index of the state in $S(s)$ after it is sorted with increasing probability of satisfying the property at time $t+1$, i.e, they are sorted with the values $p_{t+1}(s'_q)$, and $p_{t+1}(s'_0)=0$.

The sum of the objective in \eqref{eq:CPT_eq_reach2} is over the successor states. The first sum in $\Phi_{q}$ is over the actions, and the second sum in $\Phi_{q}$ computes the probability of transitioning the successor state with at least probability $p_{t+1}(s'_{q-1})$ as a function the policy.

 The problem in \eqref{eq:CPT_eq_reach2} can be viewed as maximizing the Riemann integral of the expected value, and the problem in \eqref{eq:CPT_eq_reach1} maximizes the Lebesgue integral. See the Figure 3 as an example from the MDP in Figure 4 with $\sigma(1,a)=0.3$ and $\sigma(1,a)=0.7$. Both problems will maximize the expected value, i.e, the area under the curve in Figure 3.
 
Note that the probability of satisfying the specification up to $0.2$ probability is 1, regardless of the policy we choose, as $0.2$ is the lowest probability of the successor states. Then, the probability of transitioning a state with at least $0.5$ probability of satisfying the property can be obtained by the sum of the probabilities of transitioning the state $3$ and state $4$, which is given by $\sigma(s,a)+0.4\cdot\sigma(s,b)$ in the MDP in Example 1. Similarly, the probability of transitioning to state $4$ is $0.4\cdot\sigma(s,b)$, which gives the probability of satisfying the specification with $0.9$ probability.

When $w_{+}(x)=w_{-}(x)=x$ and $u_{+}(x)=u_{-}(x)=x$, both problems in \eqref{eq:CPT_eq_reach1} and \eqref{eq:CPT_eq_reach2} can be used to maximize the expected value of satisfying the property. However, with general weighting and utility functions, we cannot use the formulation in \eqref{eq:CPT_eq_reach1}, as $w(x+y)\neq w(x)+w(y)$ in general. Therefore, with a CPT weighting and utility function, we use a modified version of \eqref{eq:CPT_eq_reach2}, because we can approximate the weighting function accurately.
 
\begin{figure}[h]

\begin{tikzpicture}
\begin{axis}[ymin=0,ymax=1.1,xmin=0,xmax=1,enlargelimits=false,
		xlabel={Probability $z$ of satisfying the property},
		ylabel={$p(z)$},
        x label style={font=\normalsize},
        y label style={font=\normalsize},width=8.5cm,height=4.0cm]
\addplot [
const plot,
fill=blue, 
fill opacity=0.2,
draw=black,
] coordinates {
(0,1) (0.2,0.58) (0.5,0.28) (0.9,0)
}
\closedcycle
;
\end{axis}

\end{tikzpicture}
\caption{The graph of the random variable with respect to the probability of satisfying the property versus probability of obtaining that value in Example 1.}
\end{figure}
\begin{example}
Consider the MDP in Figure 4 with 4 states at time $t$ with $p_{t+1}(2)=0.2$, $p_{t+1}(3)=0.5$, $p_{t+1}(4)=0.9$. The linear program that computes the maximum probability of satisfying the specification is:
\begin{align}
&\maximize\limits_{\sched(s,a),\sched(s,b)}\quad \Big(\big(\sigma(s,a)+\sigma(s,b)\big)\cdot\big(0.2\big)+\nonumber \\
 &\qquad \qquad \big(\sigma(s,a)+0.4\cdot\sigma(s,b)\big)\cdot\big(0.5-0.2\big)+\nonumber\\
 &\qquad\qquad \qquad \big(0.4\cdot\sigma(s,b)\big)\cdot\big(0.9-0.5\big)\Big)\nonumber\\
&\enskip \text{subject to}\quad \sigma(s,a)+\sigma(s,b)=1,	 \sigma(s,a)\geq 0, \sigma(s,b)\geq 0. \nonumber
\end{align}
\begin{figure}[htb!]
	\centering
\begin{minipage}[t]{0.4\textwidth}
  \centering
\begin{tikzpicture}[->,shorten >=1pt,auto,node distance=2.0cm,
                    semithick]
  \tikzstyle{every state}=[fill=none,draw=black,text=black]

  \node[initial,state]         (1) at (0, 0)  {$1$};
  \node[state] (2) at (3, 0)  {$3$};
  \node[state] (3) at (3, -1) {$4$};
   \node[state] (4) at (3, 1) {$2$};


  \path (1) edge node {$a : 1$} (2)
    (1)edge  node { } (3)
    (1)edge  node {$b : 0.6$} (4)
        ;
\node[] at (1.6,-1) {$b : 0.4$};
   
\end{tikzpicture}
  \label{fig:sub21}
\end{minipage}

\caption{An MDP with 4 states. The label $a : \gamma$ on the transitions represents that the transition happens with probability  $\gamma$, when the $a$ action is taken.}
\end{figure}

\end{example}

For general CPT weighting and utility functions, we approximate the CPT weighting function by a posynomial. Then, for a given state $s$ and horizon $t$ and the approximation function $f(p)$ with $K$ monomials, we compute a locally optimal policy by solving the following problem:

\begin{align}
&\maximize\limits_{\sched(s,a), a \in \Act} \nonumber\\ \nonumber
&\sum_{s'_{q=1} \in S(s)}^{s'_{|S(s)|}\in S(s)}\bigg(\Phi'_{q}\cdot\Big(u_+(p_{t+1}(s'_{q}))-u_+(p_{t+1}(s'_{q-1}))\Big)\bigg)\\ \nonumber
&\enskip\text{subject to} \nonumber\\
&\sum_{\act\in\Act(s)}\sched(s,\act)=1,\quad \forall\act\in\Act(s), \quad  \sched(s,\act) \geq 0,\nonumber\\
&\Phi'_{q} = \sum_{k=1}^K  c_k\cdot\Big(\sum_{\act\in\Act(s)}\sum_{s'_{m=q} \in S(s)}^{s'_{|S(s)|} \in {S(s)}}\sched(s,\act)\cdot\probmdp(s,\act,s'_{m})\Big)^{a_k}.
\label{eq:CPT_eq_reach3}
\end{align}


We highlight the differences between the optimization problems in \eqref{eq:CPT_eq_reach2} and \eqref{eq:CPT_eq_reach3}. First difference is, we replace $\Phi_{q}$ in \eqref{eq:CPT_eq_reach2} to $\Phi'_{q}$ in \eqref{eq:CPT_eq_reach3}. $\Phi_{q}$ computes the expected value of the probability of transitioning to another state with he successor state with at least probability $p_{t+1}(s'_{q-1})$ as a function the policy, and it is used in the risk-neutral measure. On the other hand, $\Phi'_{q}$ approximately computes the expected value of the probability with respect to the CPT weighting function in a CPT-based measure by approximating the CPT weighting function with $f(p)$.

The second difference is, $u_+(x)=x$ in the problem \eqref{eq:CPT_eq_reach2}. Therefore, we replace $p_{t+1}(s'_{q})$ to $u_+(p_{t+1}(s'_{q}))$. Note that the probability of satisfying the property is always between $0$ and $1$, therefore we use $u_+$ in the objective in problem  \eqref{eq:CPT_eq_reach3}.

Let $\beta \in \mathbb{R}^+$ and $g: \mathbb{R}^+ \rightarrow  \mathbb{R}^+$. The function $g(y)=y^\beta$ is concave for $0 \leq \beta \leq 1$, and convex for $\beta \geq 1$ \cite{boyd_convex_optimization}. Therefore the problem in \eqref{eq:CPT_eq_reach3} is a DC problem, and Algorithm 1 computes a locally optimal policy.

\begin{remark}
For rational $p$, the function $y^p$ can be represented by linear matrix inequalities (LMIs) \cite{alizadeh2003second}. For instance, the constraints $y^3\leq x$ and $y\geq 0$ are equivalent to
\begin{align}
\begin{bmatrix}
z & y \\
y & 1
\end{bmatrix}\geq 0 \enskip \text{and} \begin{bmatrix}
x & z \\
z & y
\end{bmatrix}\geq 0.
\end{align}
In \cite{alizadeh2003second}, it is shown that for $p=p_n/p_d>1$, we have $k(p_n,p_d)\leq \log_2{p_n}+\alpha(p_n)$, where $k(p_n,p_d)$ is the number of LMI constraints that are generated to represent $x^3\leq y$, and $\alpha(p_n)$ is a term that grows slowly compared to $\log_2$ term. Therefore, it is beneficial to use as few basis functions as possible to efficiently compute the solutions of the DC problems because we need extra variables and constraints to represent the functions $y^p$. Therefore, Chebyshev polynomials become a rather inefficient choice, as they tend to be dense polynomials with high degrees, which is required for accurate approximation. Recall that, Figure 1 shows the Chebyshev approximation diverges, when the error tolerance is set to be small.
\end{remark}

\begin{example}
Consider the MDP in Figure 4. The DC problem that computes the maximum probability of satisfying the specification, given a posynomial with $K$ basis functions,

\begin{align}
& \maximize\limits_{\sched(s,a),\sched(s,b)} \quad \sum_{k=1}^K \Bigg( c_k\cdot\bigg(\Big(\sigma(s,a)+\sigma(s,b)\Big)^{a_k}\cdot\Big(u_+(0.2)\Big)\nonumber\\
&\quad +\Big(\sigma(s,a)+0.4\cdot\sigma(s,b)\Big)^{a_k}\cdot\Big(u_+(0.5)-u_+(0.2)\Big)+\nonumber \\
&\qquad \Big(0.4\cdot\sigma(s,b)\Big)^{a_k}\cdot\Big(u_+(0.9)-u_+(0.5)\Big)\bigg)\Bigg)\nonumber\\
& \enskip\text{subject to} \quad \sigma(s,a)+\sigma(s,b)=1, \sigma(s,a)\geq 0, \sigma(s,b)\geq 0. \nonumber
\end{align}

%
%
%
%
%

We note that the objective in the above problem is a sum of DC functions and the functions in the constraints are affine functions, the above problem is a DC optimization problem and a locally optimal solution of the problem can be computed using  Algorithm 1.
\end{example}

So far, we considered formal quantitative verification of the systems, and these problems do not include cost or reward function. If we want to include cost or reward functions in a MDP to minimize the expected cost or maximize the expected reward with CPT-based measures, then objective of the optimization problem in \eqref{eq:CPT_eq_reach2} will be replaced by the following:

\begin{align}
&\maximize\limits_{\sched(s,a), a \in \Act} \nonumber\sum_{s'_{q=1} \in S(s)}^{s'_{|S(s)| }\in S(s)}\bigg(\Phi'_{q}\cdot \Big(u_+(\gamma_{s}+v_{t+1}(s'_{q}))\qquad\qquad\\
&\qquad\qquad \qquad \qquad-u_+(\gamma_{s}+p_{t+1}(v'_{q-1}))\Big)\bigg)\nonumber\\
&\gamma_{s}=\displaystyle\sum_{k=1}^K c_k \cdot\Big(\sum_{\act\in\Act(s)}\sched(s,\act)^{a_k}\cdot R_t(s,\act)\Big),
\label{eq:CPT_eq_reach5}
\end{align}
where $v_{t}(s'_{q})$ denotes the value of a state with index $q$ at time $t$.

Note that, the term $u_+(\gamma_{s}+v_{t+1}(s'_{q}))$ is a composition two convex or concave functions, which is not convex or concave in general, also that term is multiplied with a DC function $\Phi_q$. To the best of our knowledge, no general method exists to solve problems with this type of objective. But for two special cases, we can efficiently compute locally optimal solutions using CCP. If the cost or reward function is a function of state instead of state and action, then we can modify the objective function in \eqref{eq:CPT_eq_reach3} as:

\begin{align}
&\maximize\limits_{\sched(s,a), a \in \Act}\sum_{s'_{q=1} \in S(s)}^{s'_{|S(s)|}\in S(s)}\Bigg(\Phi'_q\cdot\bigg(u_+\Big(C(s)+v_{t+1}(s'_{q})\Big)\nonumber\\
&\qquad \qquad \qquad \qquad -u_+\Big(C(s)+v_{t+1}(s'_{q-1})\Big)\bigg)\Bigg). \label{eq:CPT_eq_reach6}
\end{align}


As the cost is a constant, the objective in \eqref{eq:CPT_eq_reach6} is a sum of DC functions, therefore we can compute the locally optimal solution for the case when the cost or reward function is function of a state.

The second special case we consider is when the utility functions are $u_-(x)=u_+(x)=x$. Then, adding $\gamma_s$
to the objective term in \eqref{eq:CPT_eq_reach3} will result in a formulation that computes the optimal policy for this special case.

\section{Numerical Results}

We demonstrate the proposed approach on three domains: (1) A robot in a gridworld, (2) a consensus protocol, and (3) a ride sharing example. The simulations were performed on a computer with an Intel Core i5-7200u 2.50 GHz processor and 8 GB of RAM with MOSEK \cite{andersen2012mosek} as solver and using the CVX \cite{cvx} interface.

%

\subsection{Grid world}
Consider a grid world, where states are defined as grid points on a map. An agent starts in an initial state and the objective of the agent is to reach to a given state with minimial cost. The agent can move in four directions by selecting actions (north, south, east, and west). The probability of arriving at the intended cell is $\delta$, and with probability $1-\delta$, the agent moves to a random neighboring state. The cost in each move until reaching the destination is 1 for each state. The grid world has a number of static obstacles and the agent has to avoid these obstacles as hitting an obstacle has a high cost of $M =50$. Therefore, the objective is to compute a safe (i.e., not hitting to obstacles) path that is cost efficient. In our experiments, we choose a 50 $\times$ 50 grid-world, and the gridworld MDP has 2,500 states with horizon $T=100$, $\delta=0.2$ and 300 states consisting of obstacles that the agent tries to evade. We use the weighting function $\exp(-0.5(-\ln(p))^{0.9})$ and the utility function $x^{0.88}$.

After obtaining the policies using Algorithm 1, we evaluate the policies on 500 simulation runs. The risk-neutral policy finds a shorter route (with average cost equal to 38.137 on successful runs), yet it crashes into obstacles in 41 runs. In contrast, the risk-averse policy chooses  longer routes (with average cost equal to 57.638 on successful runs), but it crashes into obstacles only in 6 runs. 

\subsection{Consensus Protocol}

This case study deals with modeling and verifying the shared coin protocol of the randomized consensus algorithm of Aspnes and Herlihy \cite{aspnes1990fast}. The shared coin protocol returns a preference 1 or 2, with a certain probability, whenever requested by a process at some point in the execution of the consensus algorithm. It implements a collective random walk parameterized by the number of processes $N$ and the constant $K > 1$ (independent of $N$). 


The first property that we want to compute is the minimal probability of finishing the process and all processes being $1$, which can be expressed as maximizing the probability states, where the execution is finished, and all coins will have the value $1$, after the process. For each benchmark instance, Figure 5 gives the number of states (\#{}states), the computation time for CPT-based measures, the minimum probability of satisfying the specification with CPT-based measure (CPT $\p$) and the actual minimum probability of satisfying the specification $\p$ in the model. We use the same weighting and utility function as in the previous example.

\begin{figure}[ht]
\centering
\begin{tabular}[b]{@{}crrlrlrr@{}}
\toprule
Parameters       & \#{}states & Time (s)  & CPT $\p$ & $ \p$ \\
\midrule
$K=2$ & $272$    & $34.49$  & $0.615$ & $0.383$                   \\  

$K=16$ & $2064$    & $384.93$  & $0.722$ & $0.484$                  \\
  
$K=64$ & $8208$    & $1961.34$ & $0.673$ & $0.498$                  \\     

\bottomrule
\end{tabular}
\caption{Results for consensus benchmark with the property of all coins having the same value.}
\end{figure}

We considered the verification of another property, where we want to compute the maximum probability of finishing the process and all coins not having the same value. Figure 6 shows the results for each instance.

\begin{figure}[ht]
\centering
\begin{tabular}[b]{@{}crrlrlrr@{}}
\toprule
Parameters       & \#{}states & Time (s)  & CPT $\p$ &  $ \p$ \\
\midrule
$K=2$ & $272$    & $41.68$  & $0.315$ & $0.108$                   \\  

$K=16$ & $2064$    & $472.19$  & $0.212$ & $0.016$                  \\

$K=64$ & $8208$    & $2953.75$  & $0.163$ & $0.002$                  \\ 

\bottomrule\end{tabular}
\caption{Results for consensus benchmark the property of all coins not having the same value.}
\end{figure}

Both examples in consensus protocol shows, the CPT-based measure tends to inflate the probability of satisfying the properties. The weighting function overestimates the small probabilities of the transition probabilities in MDPs and the utility function that we choose inflates the reachability probabilities.

\subsection{Ride Sharing}

We consider a ride sharing example, inspired by \cite{ratliff2017risk}. This case study concerns modeling the behavior of a passenger in a sequential decision-making scenario. Many ride-sharing companies set prices on their rides based on both supply of drivers and demand of passengers. Therefore, the price of a ride may fluctuate. The passengers account for the price fluctuation, which influences their behavior. 

We model the ride-sharing MDP with $S=\lbrace 0,1,2,3,4\rbrace$, where states $0,...,3$ denote the cases where the passenger did not take a ride and state $4$ represents the case when the passenger takes a ride. The price multipliers for states $0,1,2$ and $3$ are $1.0, 1.4, 1.8$ and $2.2$ respectively. $\Act=\lbrace 0, 1\rbrace$ where action $0$ is waiting, and action $1$ is taking a ride. We consider a horizon length $T=5$, and the transition matrix for action $0$ is

\begin{align*}
\mathcal{P}=\begin{bmatrix}
0.876 & 0.099 & 0.017 & 0.008\\
0.347 & 0.412 & 0.167 & 0.074\\
0.106 & 0.353 & 0.259 & 0.282\\
0.086 & 0.219 & 0.143 & 0.552
\end{bmatrix}.
\end{align*}

If action $1$ is taken, the passenger transitions to state $4$ with probability 1, which implies that a ride has been taken. We define the reward function as

\begin{align*}
R(s_t,a_t)=\begin{cases}
  \hat{R} \quad &a_t=0,\\   
  S_t - x_t(p_{base}+p_{mile}D+p_{min}T)\quad &a_t=1,
\end{cases}
\end{align*}

\noindent where $\hat{R}$ is a constant, $D$ is the distance in miles, $T$ is time in minutes, $S_t$ is a constant that decreases linearly in time, $x_t$ is the price multiplier, and $p_{base}$, $p_{mile}$, and $p_{min}$ are the base, per mile, and per min prices, respectively. We choose the prices based on Uber's Washington, DC operation\footnote{\tt\small http://uberestimate.com/prices/Washington-DC/}, and we use the same weighting function as in the previous examples and utility function $u_{+}(x)=x$. Table $1$ shows the probabilities of taking a ride at a price multiplier and time.
\begin{table}[h]
\centering
\caption{Probabilities of taking a ride with respect to the price multiplier and time.}
\label{my-label}
\begin{tabular}{|c|c|c|c|c|}
\hline
Price multiplier                 & \multirow{2}{*}{1} & \multirow{2}{*}{1.4} & \multirow{2}{*}{1.8} & \multirow{2}{*}{2.2} \\ \cline{1-1}
\multicolumn{1}{|c|}{{Time}} &                    &                      &                      &                      \\ \hline
1                               &      0.88           &                     0.25 &                  0.17    &                 0.13     \\ \hline
2                              &       0.94             &                     0.89 &               0.56       &         0.45             \\ \hline
3                                &    0.97                &                     0.83 &       0.82               &          0.78            \\ \hline
4                                &         0.99           &                     0.95 &          0.95            &            0.86         \\ \hline
     5                            &         0.99           &                     0.99 &        0.98              &        0.98              \\ \hline
\end{tabular}
\end{table}
\vspace{-0.0cm}
We note that our passenger model is relatively risk-averse, i.e, the probability of taking a ride is very high when the price multiplier is 1, and the probability decreases with increasing price multipliers. The passengers tend to take a ride with increasing time to avoid taking any further risks in case of an increase in price multiplier in the future.
\section{Conclusions}
We proposed a computational method for verification of temporal logic specifications in Markov decision processes (MDPs) with measures from cumulative prospect theory (CPT). CPT-based measures are empirically known to faithfully capture the asymmetry in the risk-averseness and risk-taking behavior of humans in decision-making.
Computation of optimal policies is impractical with CPT-based measures due to the nonlinear weighting and utility functions. The proposed method approximates the nonlinear weighting function with a difference of convex (DC) function, then computes a locally optimal policy by solving a DC problem. On the other hand, computing a policy with a CPT-based measure takes more time than computing a policy with expected-value measure, as we need to represent the DC functions as a series of linear matrix inequalities. We demonstrate the practical applicability of our approach on several scenarios. For future work, we are interested in establishing error bounds between the globally optimal CPT-value and the CPT-value that is obtained from our method in MDPs.

\bibliographystyle{abbrv}
\bibliography{literature}
\end{document}